\documentclass{elsarticle}

\usepackage{lineno,hyperref}
\modulolinenumbers[5]

\usepackage{amsmath}
\usepackage{amssymb}
\usepackage{graphicx}
\usepackage{subcaption}
\usepackage{wrapfig}
\usepackage{hyperref}
\usepackage{gensymb}
\usepackage[rflt]{floatflt}
\usepackage[utf8]{inputenc} 
\usepackage[T1]{fontenc}    
\usepackage{lmodern}
\usepackage{url}            
\usepackage{booktabs}       
\usepackage{algorithm}
\usepackage{algpseudocode}

\usepackage{color}
\definecolor{navyblue}{RGB}{55,110,205}
\definecolor{myred}{RGB}{200,30,30}

\newcommand{\real}{\ensuremath{\mathbb{R}}}

\newcommand{\eg}{\emph{e.g.}}
\newcommand{\ie}{\emph{i.e.}}
\newcommand{\kmeans}{$k$-means}

\begin{document}
\begin{frontmatter}
    \title{Autoencoder-Driven Weather Clustering for Source Estimation during Nuclear Events\tnoteref{t1}}
    \tnotetext[t1]{© 2018. This manuscript version is made available under the CC-BY-NC-ND 4.0 license \url{http://creativecommons.org/licenses/by-nc-nd/4.0/}}
    \author[iit]{Iraklis A. Klampanos}
    \author[iit]{Athanasios Davvetas}
    \author[inrastes]{Spyros Andronopoulos}
    \author[inrastes]{Charalambos Pappas}
    \author[inrastes]{Andreas Ikonomopoulos}
    \author[iit]{Vangelis Karkaletsis}
    \address[iit]{Institute of Informatics \& Telecommunications, National Centre for Scientific Research ``Demokritos'', Greece}
    \address[inrastes]{Institute of Nuclear \& Radiological Sciences and Technology, Energy \& Safety, National Centre for Scientific Research ``Demokritos'', Greece}

\begin{abstract}
Emergency response applications for nuclear or radiological events can be significantly improved via deep feature learning due its ability to capture the inherent complexity of the data involved. In this paper we present a novel methodology for rapid source estimation during radiological releases based on deep feature extraction and weather clustering. Atmospheric dispersions are then calculated based on identified predominant weather patterns and are matched against simulated incidents indicated by radiation readings on the ground. We evaluate the accuracy of our methods over multiple years of weather reanalysis data in the European region.  We juxtapose these results with deep classification convolution networks and discuss advantages and disadvantages. We find that deep autoencoder configurations can lead to accurate-enough origin estimation to complement existing systems, while allowing for rapid initial response. 
\end{abstract} 

\end{frontmatter}

\section{Introduction}
\label{sec:introduction}

In the field of atmospheric dispersion modelling and its applications for supporting decision making during events of atmospheric releases of hazardous substances (\eg{} radioactive material), \emph{inverse source-term estimation} and \emph{source inversion} refer to computational methods aiming at estimating the location and/or the emitted quantities of the hazardous material using both observations (readings on the ground) and results of dispersion models. Such methods are typically used when the presence of a hazardous substance above the background levels in the air is detected by an existing monitoring network, while its origin is unknown.

The most characteristic example of a real case involving radioactive elements that have been detected before the release was officially announced is the Chernobyl Nuclear Power Plant accident\footnote{\url{http://www.irsn.fr/EN/publications/thematic-safety/chernobyl/}}. The Acerinox (or Algeciras) incident is another example of an unknown radioactive release that was traced back after radioactivity levels higher than the background had been observed at very long distances from the release location \cite{MMS12201}.

Depending on various factors, such as the spatial resolution desired, traditional inverse modelling can be computationally expensive, and therefore time-consuming, rendering its application problematic when timing is critical. In addition, atmospheric dispersion models (\eg{} HYSPLIT \cite{Draxler1999,Stein2015}) are complex pieces of software that require expert training and case-by-case application. In this paper we present a novel, complementary approach based on data analytics and deep learning. Our goals are to effectively transfer the computational bulk before the time of such an event taking place, create reusable data by-products of value and complement existing emergency response methodologies. While the computational bulk, involving the training of deep feature learning models, is inherently time-consuming, it leads to rapid (in the order of a few seconds) initial estimates during events through the reusability of the data and models generated.

Focusing on the European region, we cluster re-analysis weather data in order to derive weather circulation patterns, which affect plume dispersion. We then calculate plume dispersions for a number of European nuclear power plant facilities of interest, based on representative cluster descriptors. Last, we match previously unseen weather to the weather patterns we have learned and rank the nuclear facilities according to how well their plume dispersion for the closest weather patterns match current hypothetical radiation measurements.

This work combines data analytical and machine learning methods and large weather and atmospheric dispersion datasets and models in a single framework for rapid source estimation.

We make the following contributions:
\begin{enumerate}
    \item we propose and evaluate a novel data-driven methodology for release origin estimation;
    \item we evaluate a series of autoencoder configurations, followed by \kmeans{} clustering applied on weather re-analysis data;
    \item we juxtapose our method with deep convolution networks for classification, discussing their respective advantages and disadvantages; and
    \item we present an application prototype for the rapid estimation of release origin and its implementation.
\end{enumerate}

In the following section we provide a succinct discussion of related approaches, technologies and methods. In Section~\ref{sec:methodology} we present the proposed methodology and rationale, while in Section~\ref{sec:results_and_discussion} we provide a discussion of our evaluation methodology and discuss results and findings. In Section \ref{sec:app_prot} we present a pilot application showcasing the proposed approach. In Section~\ref{sec:conclusions} we conclude the paper and discuss directions for future work, while in Section \ref{sec:ack} we point to software resources to encourage cross-examination and replication.

\section{Related Work}
\label{sec:background}
This study combines algorithms and methods borrowed from a number of disciplines, such as machine learning, weather circulation clustering, atmospheric dispersion and weather modelling. An overview of relevant work is provided below. 

\subsection{Discovering Weather Patterns}
\label{sub:weather_clustering}
Over the last decades there have been several studies attempting to automatically discover weather patterns via unsupervised hierarchical and iterative clustering \cite{Huth2008}, as part of synoptic climatology \cite{Yarnal2001}. Many of these studies have attempted to establish statistically robust representations based on the calculation of principal components in, what this body of literature refers to as, S- and T-modes. S-mode refers to the typical application of principal component analysis (PCA) per weather sample used for feature reduction, while T-mode refers to the application of PCA per grid point or feature. (In our evaluation, in Section \ref{sec:results_and_discussion}, we refer to the T-mode also as PCA$^{T}$, to indicate cases where PCA has been applied on the ``transposed'' form of the data samples.) Indicatively, Huth \cite{Huth1996} summarises and compares correlation, sums-of-squares, agglomerative hierarchical, PCA, \kmeans{} and hierarchical agglomerative clustering algorithms. These algorithms were executed on geopotential height (GHT) at a pressure level of 700$hPa$ and were evaluated using internal metrics, such as consistency and robustness. According to Huth, there is no clear winner for clustering weather patterns, however T-mode PCA appears to produce clusters that resemble manually identified weather patterns more accurately.  Methods based on more recent advances in neural networks, employing self-organising maps, have also been reported \cite{Cavazos2000,Hewitson2002}. 

Other classification and clustering approaches tailored to specific applications or geographical regions have been published. Teixeira de Lima and Stephany \cite{TeixeiradeLima2013} evaluate a multitude of weather variables for extreme weather classification. Straus et.al. \cite{Straus2007} use a modified version of \kmeans{} clustering for the pressure level of 200$hPa$ and zonal winds to discover and analyse winter circulation regimes over the Pacific-North American region.
Hsu and Cheng \cite{Hsu2016} evaluate daily average surface wind measurements to discover weather patterns affecting air pollution in western Taiwan. Focusing on the urban heat island, the phenomenon in which an urban area is significantly warmer than its surrounding rural areas due to human activity, Hoffmann and Schl\"unzen \cite{Hoffmann2013} study \kmeans{} on a number of variables such as GHT, relative humidity, vorticity, and others. Bannayan et.al. use temperature, precipitation and solar radiation in a k-nearest neighbour approach for real-time prediction of daily weather data \cite{Bannayan2008}. Extracting information out of weather patterns has also been reported in cases of fire spread modelling by Duane et.al. \cite{Duane2016}. 
Al-Alawi et.al employ PCA for feature reduction, as part of a combination of principal component regression and feed-forward neural networks, for the prediction of ozone concentration \cite{Al-Alawi2008}. 

Far from being exhaustive, the above studies are indicative of the multitude of applications, clustering and classification approaches reported in the wider area of environmental modelling.
While part of our study involves discovering useful weather patterns, our application of these clusters is specific to capturing the conditions leading to similar plume dispersions. Our experiments, reported in Section \ref{sec:results_and_discussion}, focus on GHT as a feature predictive of circulation patterns and consequently, plume dispersion.

\subsection{Autoencoders for Feature Reduction} 
\label{sub:autoencoders_for_feature_reduction}
An autoencoder \cite{LeCun1998,Bengio2009,Goodfellow2016} is an unsupervised feed-forward neural network designed to approximate the identity function, \ie{} one that attempts to learn a function $h(x) = \hat{x} \approx x$, where $x$ and $\hat{x}$ denote the input and output vectors respectively. Post-training, applications typically disregard the output of autoencoders, instead making use of the activation values of the hidden layers. These constitute latent representations of the input. The activations of simple, single-layer auto-encoders have been shown to be equivalent to principal components \cite{Bourlard1988}.

In its simplest form, when there is a single hidden layer and the number of hidden units equals the number of inputs, the auto-encoder is too successful in replicating the input, leading to overfitting. Various methods have been suggested to avoid overfitting, \eg{} having fewer hidden than input units, enforcing activation sparsity or introducing noise which the auto-encoder learns to compensate for \cite{Vincent2008,Vincent2010}. An alternative approach is to use deeper configurations of stacked autoencoders, where inner layers encode and decode previously encoded vectors. Encodings generated by stacked autoencoders can capture deeper statistical representations of the input data.

Autoencoders have been augmented by different types of deep neural networks, such as convolutional networks, or \textit{convnets}. Convnets \cite{lecun-98,Goodfellow2016} are designed to discover multi-dimensional patterns of varying sizes and have been used, stand-alone or as part of more complex configurations, primarily in classification-based image recognition. 
Convolutional autoencoders are formed by stacking convolutional layers, fully connected layers and de-convolutional layers in a single configuration in order to capture feature hierarchies in the input space \cite{Masci2011}.

In this study we evaluate simple, stacked and convolutional denoising autoencoders \cite{Masci2011} to reach smaller and statistically robust representations of weather variables. We then use latent representations to discover weather patterns in Europe using the \kmeans{} algorithm.

\section{Methodology and Rationale} 
\label{sec:methodology}

\begin{figure}
	\includegraphics[width=\textwidth]{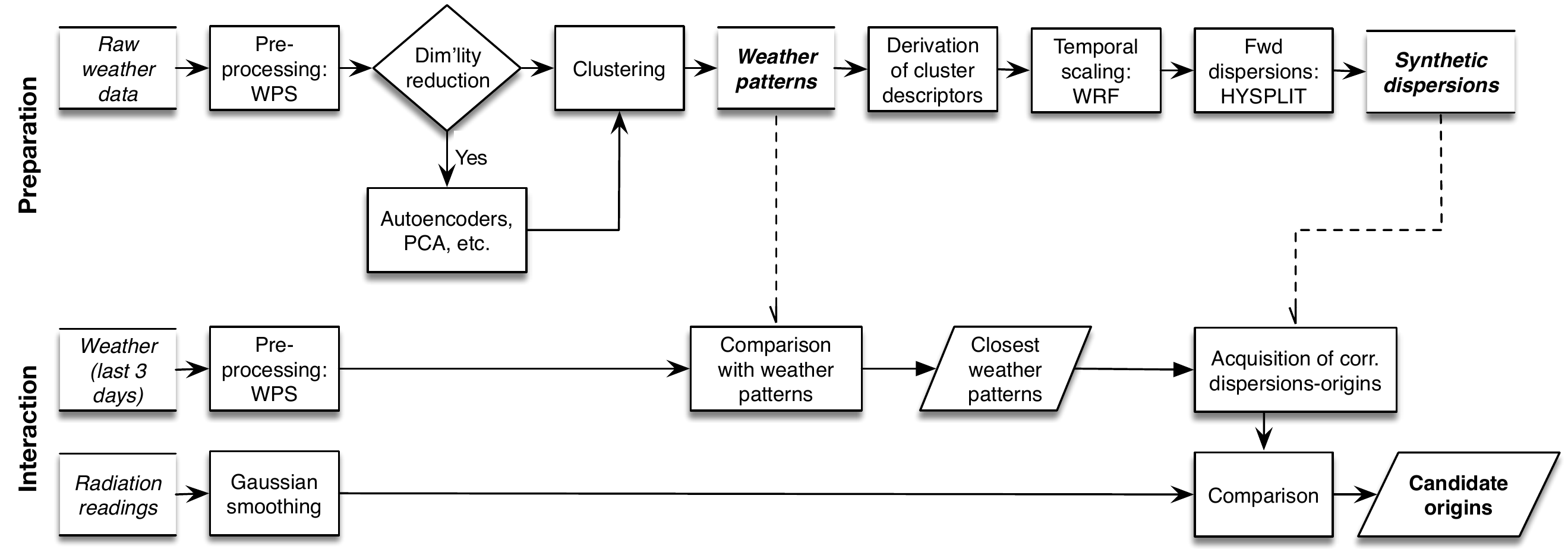}
    \caption{Methodology overview. Computationally-expensive operations take place during the preparation phase, allowing the interaction phase to produce rapid origin estimations.}
	\label{fig:methodology}
\end{figure}

Our methodology comprises two main steps: (1) the derivation of weather patterns to be used for driving reference plume dispersion patterns, and (2) the subsequent estimation of the release origin. An overview of our methodology is shown in Fig. \ref{fig:methodology}.

\subsection{Weather Re-analysis Data and Features} 
\label{sub:weather_reanalysis_data_and_features}
To discover weather patterns we use weather re-analysis data. This is gridded data that originates from optimal combinations of Numerical Weather Prediction (NWP) data and observations, therefore representing the best available description of past atmospheric conditions. Weather re-analysis data cover a sufficiently large period (40 years) and are of fine enough spatial (less than 1$\degree$) and temporal (up to 6hr) resolution over the region of interest, \ie{} the European continent. This dataset is freely available online as part of the ERA-Interim project \cite{Dee2011} and is provided by the European Centre for Medium range Weather Forecast (ECMWF)\footnote{\url{http://www.ecmwf.int/en/research/climate-reanalysis/era-interim}}.

For this study we used global ERA-Interim data covering 11 years (1986-1993 and 1996-1998) with a 6hr temporal resolution provisioned by NCAR in the GRIB format\footnote{\url{http://rda.ucar.edu/datasets/ds627.0/}}. The resolution of the original data is $\approx0.7\degree\times0.7\degree$. These data contain several atmospheric variables expressed across $37$ vertical pressure levels ranging from 1$hPa$ to 1000$hPa$ -- pressure levels can be seen as a more robust altitude representation.
In this study we focus on the geopotential height (GHT) variable, a ``gravity-adjusted'' height, and more specifically to either GHT\@700$hPa$ or a combination of GHT@500, 700 and 900$hPa$. GHT has been shown to be predictive of weather circulation patterns \cite{Huth1996,Huth2008}.
Raw weather reanalysis data is pre-processed via the Weather Research and Forecasting (WRF) \cite{Michalakes2004} preprocessor WPS to define meteorological data on a cartesian domain that covers Europe, with a spatial resolution of $64\times64$ cells of $75km\times75km$ in the west-east and south-north directions. In the vertical direction, we retained the original pressure levels. 

\begin{figure}
    \centering
    \includegraphics[width=.85\textwidth]{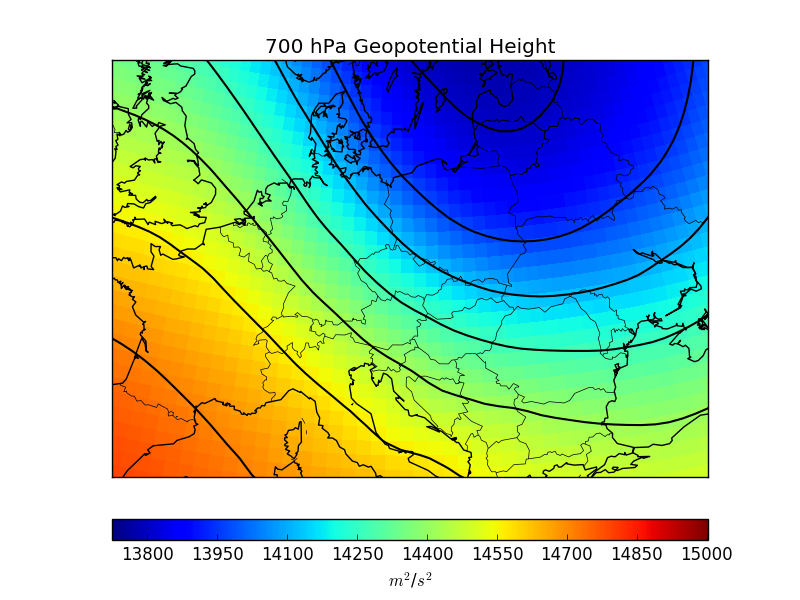}
    \caption{Sample of the GHT variable at a pressure level of 700$hPa$ over part of Europe.}
	\label{fig:ght-example}
\end{figure}

The resulting data used to derive weather patterns are either 2D grids of 64$\times$64 cells,  or 3D grids of 64$\times$64$\times$3 cells, i.e. for the three atmospheric pressure levels of interest mentioned above. An example of GHT at the 700$hPA$ pressure level is shown in Fig. \ref{fig:ght-example}.
The weather patterns discovered as a result of feature extraction and analysis are ``downscaled'' temporally using WRF, resulting to hourly data of the same spatial dimensionality as before. These data are then used for the calculation of forward dispersions. The resulting atmospheric dispersion grid has a size of 500$\times$500 cells, corresponding to approx. 10km$\times$10km, and a temporal resolution of 1hr.

\subsection{Dispersion and Inverse Models} 
\label{sub:dispersion_models}
Atmospheric dispersion models are computational codes that simulate the processes of transport and diffusion of air pollutants, as well as other physical processes that occur during dispersion, such as deposition on the ground and transformations (chemical reactions, radioactive decay, etc.). Dispersion model calculations are based on meteorological data, such as these processed by WRF, described above.

Dispersion models can operate in ``forward'' in time (prognostic) mode when the pollutant emission locations and the related emitted quantities (as functions of time) are known. They can also be used in ``inverse'' computations, where in combination with available measured pollutant concentrations, can estimate unknown pollutant emission locations and emission rates. There are several types of atmospheric dispersion models. In this work, which involves dispersions in long spatial ranges over complex topography and high meteorological variability, we use the so-called ``Lagrangian'' models. These models simulate the emission of pollutants by a series of virtual ``puffs'' or particles that are transported by the wind field and diffused by the atmospheric turbulence. Air concentration of pollutants and deposition on the ground are computed on a grid defined independently of the meteorological data grid.

\begin{figure}
\centering 
\includegraphics[width=\textwidth]{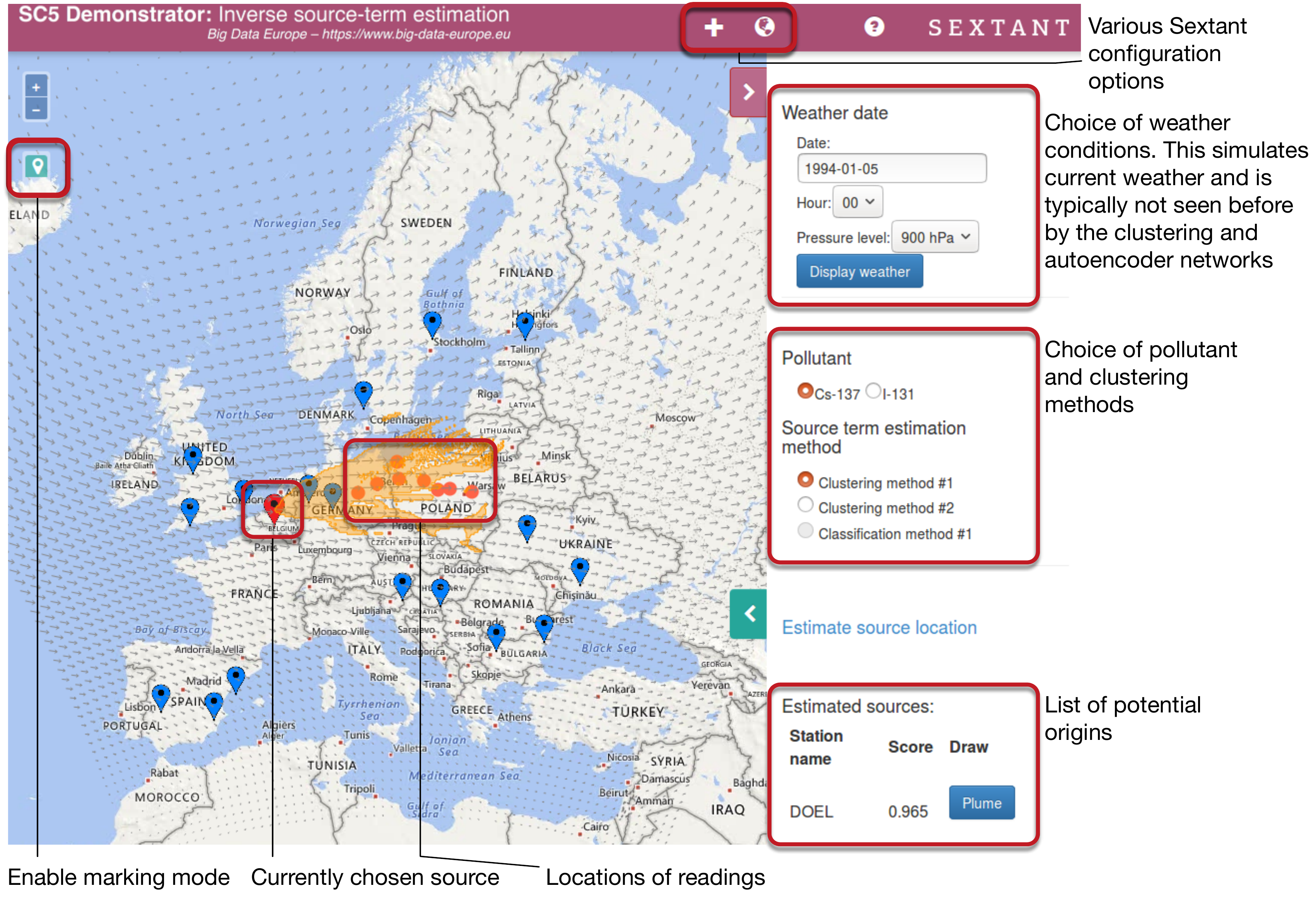}
\caption{The user interface of the application prototype. The dots indicate the locations of radiation readings while the arrows represent the current wind direction. The main component of the user interface is the map, which acts as the base layer where additional information is stacked on top of. On the right-hand side there is the control panel where the user can select different configurations of the source estimation procedure as well as the most probable sources allowing for the visualisation of their respective plumes.}
\label{fig:screenshot}
\end{figure}

In our experiments we make use of the widely used NOAA HYSPLIT atmospheric dispersion model \cite{Draxler1999,Stein2015}, which works well with data produced by WRF. Other dispersion models, such as the DIPCOT model \cite{Andronopoulos2010,Tsiouri2010}, can also be used. A visual of a 2D dispersion as displayed by our application is shown in Fig. \ref{fig:screenshot}.

\subsection{Weather Patterns and Cluster Descriptors for Plume Dispersion} 
\label{sub:deriving_weather_patterns_for_dispersion}

\emph{\kmeans{} clustering:} Data clustering \cite{Jain1999,Bishop2006} represents a broad field dealing with the unsupervised discovery of groupings in multidimensional data based on their statistical similarity. A multitude of clustering algorithms has been proposed in the literature, many of which being used in operational settings.
In this work we derive weather patterns via \kmeans{} clustering. The calculation of these patterns and their associated atmospheric dispersions for a number of nuclear facilities of interest are used for subsequent origin estimation.

\kmeans{}, essentially an expectation-maximization (E-M) type of algorithm, is one of the most widely-used algorithms for data clustering. The algorithm is initialised by $k$ typically random cluster centroids $\{\mu_i\}$ and its goal is to assign data points to a set of $k$ corresponding sets $C=\{c_i\}$ so as to optimise an objective function;
During the expectation step each data point is assigned to the closest cluster based on its distance from the clusters' centroids. During the maximization step the centroids $\{\mu_i\}$ get updated to reflect the new assignments. The algorithm stops when either there are no changes in the cluster assignments, or when a predefined threshold on the acceptable number of iterations has been reached. \kmeans{} is guaranteed to eventually converge, however not necessarily towards the global minimum.

\emph{Autoencoders for dimensionality reduction:} Working with long sequences of dense high-dimensional vectors can be impractical. More importantly, weather exhibits spatiotemporal patterns which may not be immediately exploitable by dealing with the raw data. We perform dimensionality reduction on the raw weather data using autoencoders in order to obtain reduced, statistically robust summaries. We then derive weather patterns by feeding the encoded versions of weather snapshots to \kmeans.

Several dimensionality reduction techniques can be applied on weather data, such as PCA. We base our choice to use autoencoders on two premises (1) a shallow, linear activation auto-encoder is equivalent to PCA therefore covering our evaluation requirements and (2) given enough training, stacked autoencoders may lead to more robust representations spanning multiple variables and therefore lead to better quality clustering. In Section~\ref{sec:results_and_discussion} we evaluate different configurations of autoencoders and discuss their effectiveness further.

\emph{Cluster descriptors:} The clusters derived are hypothesised to consist of weather snapshots representative of specific weather patterns observed in Europe and are described by cluster descriptors or summaries. A clustering outcome $C$ is a set of clusters $\{c_i \in C | c_i=[w_t \in \real{}^{xyzm}]\}$, where $w_t$ is a value at time $t$. Variables $x$, $y$ and $z$ indicate the geographical location and geopotential height respectively, and $m$ denotes the physical variable, \eg{} the geopotential, wind velocity, etc. By default \kmeans{} uses centroids as descriptors. Even though this is sensible for fitting additional weather snapshots to the model, it is not useful for producing synthetic atmospheric dispersions as it does not contain any temporal information.
Taking into account errors introduced by unsupervised clustering as well as the dispersion models requiring time information, cluster descriptors should achieve two goals: they should reflect dominant weather conditions within the clusters and lend themselves to temporal and physical interpretation.

In this work we derive cluster descriptors, which are used by dispersion models as sequences of representative weather snapshots, following two approaches (also outlined in Algorithm \ref{alg:descr}):

\begin{algorithm}
    \caption{Pseudocode for the \textit{km}$^2$ and \textit{dense} cluster descriptors}
\label{alg:descr}
\begin{algorithmic}[1]
    \Procedure{km$^2$}{$W$} \Comment Lists of weather snapshots
    \State \textbf{return} $k$-means($W$, $k=13$).centroids
    \EndProcedure
    \State

    \Procedure{Dense}{$W, T$} \Comment{Lists of weather snapshots and their times}
    \State window $\gets$ 13 \Comment Number of 6-hourly snapshots
    \For{$t_i \in T$} \Comment Flatten all times to a single year
    \State $t_i$.year $\gets 0$
    \EndFor
    \State $T' \gets $ $[$Mean($[t_j]$ for $t_j \in T$ and $t_j$s sharing date and month$)]$
    \State density $\gets$ GaussianKernelDensity($t' \in T'$) \Comment{For some bandwidth}
    \State maxIndex $\gets$ argmax density
    \State startIndex $\gets$ maxIndex $-$ window$/ 2$
    \State endIndex $\gets$ maxIndex $+$ window$/ 2$
    \State descriptor $\gets [w_j \text{ if } w_j \in W; \text{ else } Null]$ for $j\in [\text{startIndex},\text{endIndex})$
    \For{$x \in \text{descriptor}$} \Comment Pad missing snapshots by copying adjacent
    \If{$w_j$ = Null}
    \State $w_j \gets w_{j+1}$ or $w_{j-1}$
    \EndIf
    \EndFor
    \State \textbf{return} descriptor
\EndProcedure
\end{algorithmic}
\end{algorithm}

\begin{enumerate}
    \item The first approach is to derive these snapshots by a second application of \kmeans{} within each cluster, taking $k$ to be the number of desired snapshots (in this case $k=13$ -- since we work on 3-day intervals we take 12 6-hourly snapshots plus 1 snapshot to inform the dispersion model of the end time of the dispersion run).
The centroids of the resulting clusters form the original cluster's descriptor. This approach should be able to capture all weather trends within each cluster effectively, however it does not retail meaningful temporal information. We refer to this cluster descriptor approach as \emph{km$^2$}: the \kmeans{} of the \kmeans{} result.
    \item A second approach is based on temporal densities within each cluster. After collapsing the weather snapshots of each cluster from multiple years into a single-year period, we calculate its temporal density. (The collapsing into a single-year period is to enforce time continuity and therefore improve the subsequent calculation of the dispersion pattern.) We then choose the weather snapshot corresponding to the time maximising density and select a temporally continuous period around it for as many items as we require the descriptor to have. If the cluster has more than one snapshots at this offset from the beginning of the year (originally coming from multiple years) the mean weather is taken into account. If the cluster lacks continuous snapshots we use copies of neighbouring ones.  Whereas cluster descriptors created by this approach have a more sound temporal and physical interpretation, they fail to adequately represent poorly formed clusters. We refer to these descriptors as density-based or \emph{dense}.
\end{enumerate}

After weather snapshots have been clustered and weather patterns have emerged, we simulate hypothetical releases from 20 European nuclear facilities. Using the HYSPLIT atmospheric dispersion model \cite{Draxler1999,Stein2015} we calculate hypothetical dispersions of radioactive material into the atmosphere for each of these locations. For the purposes of this application we assume that the emitted particles have a fixed atomic weight. These atmospheric dispersions have a finer geospatial resolution and form the basis for origin estimation.

\subsection{Source Estimation} 
\label{sub:estimation_of_release_origin}
To estimate the release origin we follow a two-step matching procedure. After an event has taken place we first identify the weather pattern that best represents
the current weather.
For this calculation several similarity (or distance) metrics can be used. Treating this matching as an additional \kmeans{} assignment step, we choose the cluster which minimises the Euclidean distance between the observed weather vector $w$ and the centroid of the cluster in question.

Based on our choice of a representative weather pattern we then consider the dispersions previously calculated for our set of fixed candidate release origins. The likelihood of a location being the release origin can be calculated by comparing the release distributions of the cluster-based dispersions against hypothetical detection readings.

A hypothetical event is detected by a sequence of readings at certain locations. We model these point readings as a discrete probability distribution across the geographical area of study $r(x,y):\real$ (after dividing each reading by their total sum). This evaluates to a reading at all locations where radiation has been detected and to $0$ at all other locations. The readings distribution is compared against a number of dispersion distributions given a weather pattern and a point of origin.

Since $r(x,y)$ is typically sparse we pass it through an isotropic 2D Gaussian smoothing filter obtaining $R = r_g(x,y)$. Based on the assumption that cells neighbouring readings are likely to also be contaminated, albeit with a decreasing probability, applying such a filter increases the number of cells with non-zero radiation values and therefore the likelihood to obtain meaningful positive matches between readings and dispersions.

Finally, we take the inverse order of the dispersions corresponding to the weather pattern in question $D$ according to their distance from the detection distribution $R$. Due to their sparsity, we compare the readings and dispersion vectors using the cosine distance metric.

\begin{figure}
\includegraphics[width=\textwidth]{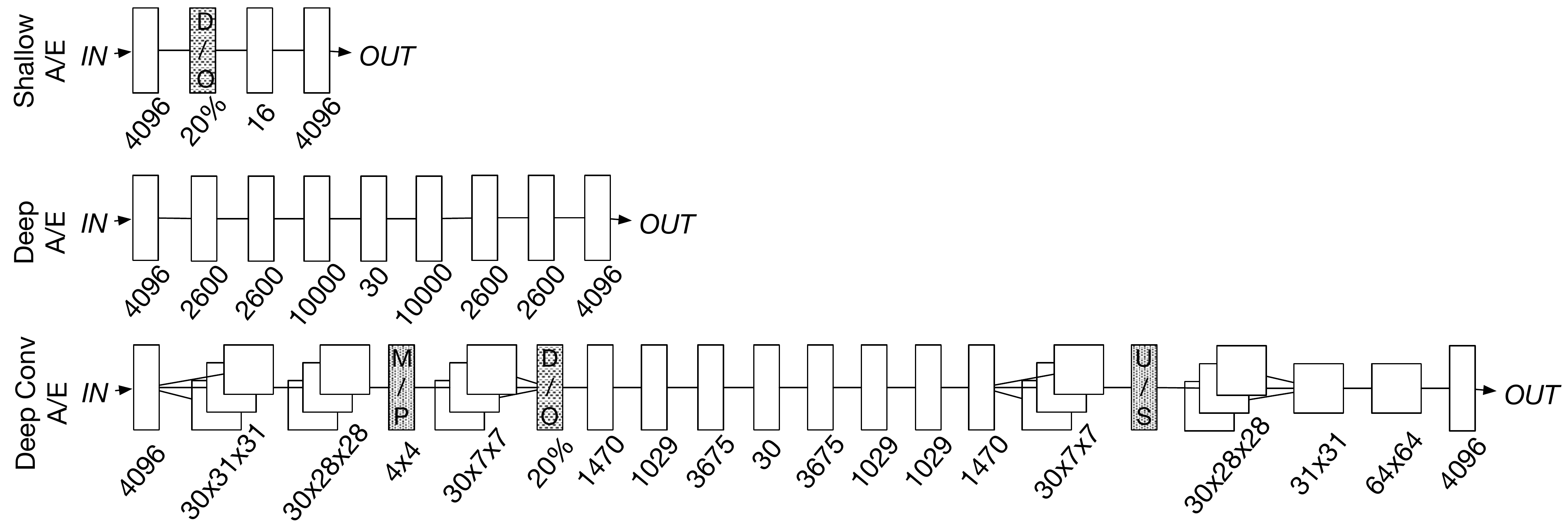}
\caption{The autoencoders implemented and evaluated. The first autoencoder (\emph{Shallow A/E}) is designed to approximate PCA. A deeper autoencoder (\emph{Deep A/E}) attempts to better discover latent weather structure. A convolutional autoencoder (\emph{Deep Conv A/E}) aims to expose yet better features by exploiting spatial neighbourhoods in the GHT distribution. An additional convolutional autoencoder (\emph{Deep MC Conv A/E}) is also implemented and reported in this study. It is not shown for brevity as its configuration is identical to \emph{Convolutional AE}, other than it operates on three variables, GHT at 500, 700 and 900hPA, provided as channels. D/O are dropout layers which introduce noise by randomly setting to zero 20\% of the features. In the case of the Deep A/E, a drop-out of 20\% was only used during the layer-wise training. M/P indicates maxpool, while U/S indicates upscaling. All activations used were ReLUs, apart from the first and last layers, where linear activations were used. Weights were updated using stochastic gradient descent (SGD) with Nesterov momentum (0.9), a minibatch size of 265 and training lasted 200 epochs. Inputs were scaled so that they had a mean $=0$ and variance $\approx 1$.}
\label{fig:autoencoders}
\end{figure}

\section{Evaluation and Discussion}
\label{sec:results_and_discussion}
The autoencoders that drive weather clustering were trained using 11 years of 6-hourly weather snapshots (1986-1993 and 1996-1998\footnote{This split in time is a result of incrementally upscaling our experiment with additional years. The dimensionality reduction algorithms evaluated do not take temporal continuity into account as they operate on discrete weather snapshots.}). Weather data used for training and evaluation were represented by a 64$\times$64 grid. In order to obtain more samples to train with we introduced $\pm10\%$ element-wise random noise to the weather data and increased the number of samples five-fold. This resulted in a total of 96,432 training samples.

Our cluster-based system was evaluated using 3-day dispersions spanning 2 years of European weather (1994-1995), for a choice of 20 nuclear facilities, resulting in 4,480 samples. We increased the evaluation sample size five-fold by introducing the same type of element-wise random noise as with the weather samples.

For creating training and evaluation datasets we approximated radiation readings by randomly choosing 10 and 30 geographical points -- identifiable by (latitude, longitude) pairs -- from each sample's dispersion pattern. This simulates receiving readings from arbitrary sensors on the ground and assesses how the overall accuracy changes with the amount of available information.
In real situations readings are not expected to be randomly scattered across the area under a plume's path of travel. This choice for training and evaluation was deliberate as it makes no assumptions regarding the actual locations of radiation detectors\footnote{European countries are equipped with stationary networks of radiation detectors.}, while it is more readily generalisable to different use-cases, such as chemical and other releases from unknown potential origins.

\begin{table}
    \caption{Results for 10 simulated reading locations, reporting accuracy as a function of the dimensionality reduction configuration and the choice of the cluster descriptor. Raw \kmeans{} (0) reports on accuracy when no feature selection is used, \ie{} the input data are used in raw form, Shallow A/E (1) uses a shallow autoencoder with one hidden layer, Deep A/E (2) employs a deep autoencoder, Deep Conv A/E (3) reports on a deep convolutional autoencoder, Deep MC Conv A/E (4) reports on a multi-channel, deep, convolutional autoencoder, PCA (5) reports on PCA feature reduction into 16 principal components and PCA$^T$ (6) reports on PCA$^{T}$ in order to acquire representative weather snapshots across the samples, followed by typical PCA feature reduction. Accuracy@1 considers only the top estimated origin, while Accuracy@3 considers the top 3 results. These results are obtained by applying the cosine distance between hypothetical readings and dispersion pattens in vector form. Details on layers and layer sizes of the autoencoders evaluated can be found in Fig. \ref{fig:autoencoders}}.
\centering
{
\begin{tabular}{|c|c|c|c|c|c|}
\cline{3-6}
\multicolumn{1}{c}{} &  & \multicolumn{2}{c|}{\emph{Accuracy@1 (\%)}} & \multicolumn{2}{c|}{\emph{Accuracy@3 (\%)}}\tabularnewline
\hline
\emph{Idx} & \emph{Configuration} & \emph{km$^{2}$} & \emph{density} & \emph{km$^{2}$} & \emph{density}\tabularnewline
\hline
\hline
0 & Raw \emph{k-}means & 28.745  & 29.049  & 58.415  & 58.107 \tabularnewline
\hline
1 & Shallow A/E & \textbf{30.973} & 29.620 & \textbf{60.910} & 60.075\tabularnewline
\hline
2 & Deep A/E & 30.084  & \textbf{30.303}  & 59.919  & \textbf{61.156} \tabularnewline
\hline
3 & Deep Conv A/E & 30.544  & 28.790  & 60.303  & 58.325 \tabularnewline
\hline
4 & Deep MC Conv A/E & 26.633  & 30.053  & 55.892  & 60.178 \tabularnewline
\hline
5 & PCA (components=16) & 29.732  & 29.093  & 59.151  & 59.473 \tabularnewline
\hline
6 & PCA$^{T}$ & 29.357  & 25.066  & 58.419  & 56.843\tabularnewline
\hline
\end{tabular}
}
\label{tbl:ae-results10}
\end{table}

\begin{table}
    \caption{Results for 30 simulated reading locations, reporting accuracy as a function of the dimensionality reduction configuration and the choice of the cluster descriptor. Dimensionality reduction configurations as of Table \ref{tbl:ae-results10}. These results demonstrate that more readings lead to significantly higher accuracy.}
\centering
{
\begin{tabular}{|c|c|c|c|c|c|}
\cline{3-6}
\multicolumn{1}{c}{} &  & \multicolumn{2}{c|}{\emph{Accuracy@1 (\%)}} & \multicolumn{2}{c|}{\emph{Accuracy@3 (\%)}}\tabularnewline
\hline
\emph{Idx} & \emph{Configuration} & \emph{km$^{2}$} & \emph{density} & \emph{km$^{2}$} & \emph{density}\tabularnewline
\hline
\hline
0 & Raw \emph{k-}means & 37.732  & 38.477  & 69.906  & 69.562\tabularnewline
\hline
1 & Shallow A/E & \textbf{40.883} & 38.433 & \textbf{73.906} & 72.263\tabularnewline
\hline
2 & Deep A/E & 40.616  & 39.267  & 72.696  & 72.705\tabularnewline
\hline
3 & Deep Conv A/E & 39.906  & 38.901  & 72.129  & 70.995\tabularnewline
\hline
4 & Deep MC Conv A/E & 35.861  & \textbf{41.638}  & 67.500  & \textbf{74.513}\tabularnewline
\hline
5 & PCA (components=16) & 40.165  & 37.553  & 72.245  & 71.348\tabularnewline
\hline
6 & PCA$^{T}$ & 39.397  & 32.808  & 70.160  & 68.558\tabularnewline
\hline
\end{tabular}
}
\label{tbl:ae-results30}
\end{table}

The quality of the clustering is crucial to the performance of this application as a whole, which in turn depends on the quality of the latent features exposed by the autoencoder. In this paper we study denoising autoencoders of different depths and complexities, shown in Fig. \ref{fig:autoencoders} -- the shallow autoencoder approximates PCA \cite{Bourlard1988}, while deeper configurations are expected to lead to more performant clustering solutions. Further fine-tuning the networks in order to improve the clustering outcome, \eg{} by using the DEC algorithm \cite{Xie2016}, could further increase performance and it is left as future work.

Cluster analysis on both raw and encoded weather data was inconclusive regarding the optimal number of clusters to be used, $k$. Based on cluster consistency scores and subjective expert opinions regarding observed weather patterns in Europe we set $k$=15.

With the clustering algorithm and the choice of atmospheric variables remaining constant, the accuracy of our system depends on the following parameters: (1) the dimensionality reduction approach applied -- in this case on the configuration of the autoencoder; (2) the choice of the distance (or similarity) metric between synthetic dispersions and radiation readings; (3) the choice of cluster descriptors for deriving synthetic dispersions; and (4) the number of available readings.

Tables \ref{tbl:ae-results10} and \ref{tbl:ae-results30} summarise results obtained for different choices of dimensionality reduction algorithms and cluster descriptors for 10 and 30 readings respectively. We have experimented both with the cosine
and the correlation
distances between vectors of readings and synthetic dispersions, with the former being consistently better. Euclidean-based distances are not performant due to the sparsity of the vectors involved. Here, we report results obtained via the cosine distance metric.

In Tables \ref{tbl:ae-results10} and \ref{tbl:ae-results30} we observe that the km$^2$ descriptor performs best with the shallow autoencoder (\emph{Shallow A/E}). This can be attributed to the fact that the km$^2$ descriptor effectively counters cluster inconsistencies averaging over often meaningful sub-clusters. However, the sequence in which the averages are produced is indeterminable, yet time is meaningful to the dispersion model. For the multi-channel convolutional autoencoder density-based descriptors work best. Because \emph{Deep MC Conv A/E} leads to better clustering outcomes, temporal density better represents weather clusters.

\begin{figure}
	\centering
 	\includegraphics[width=.8\textwidth]{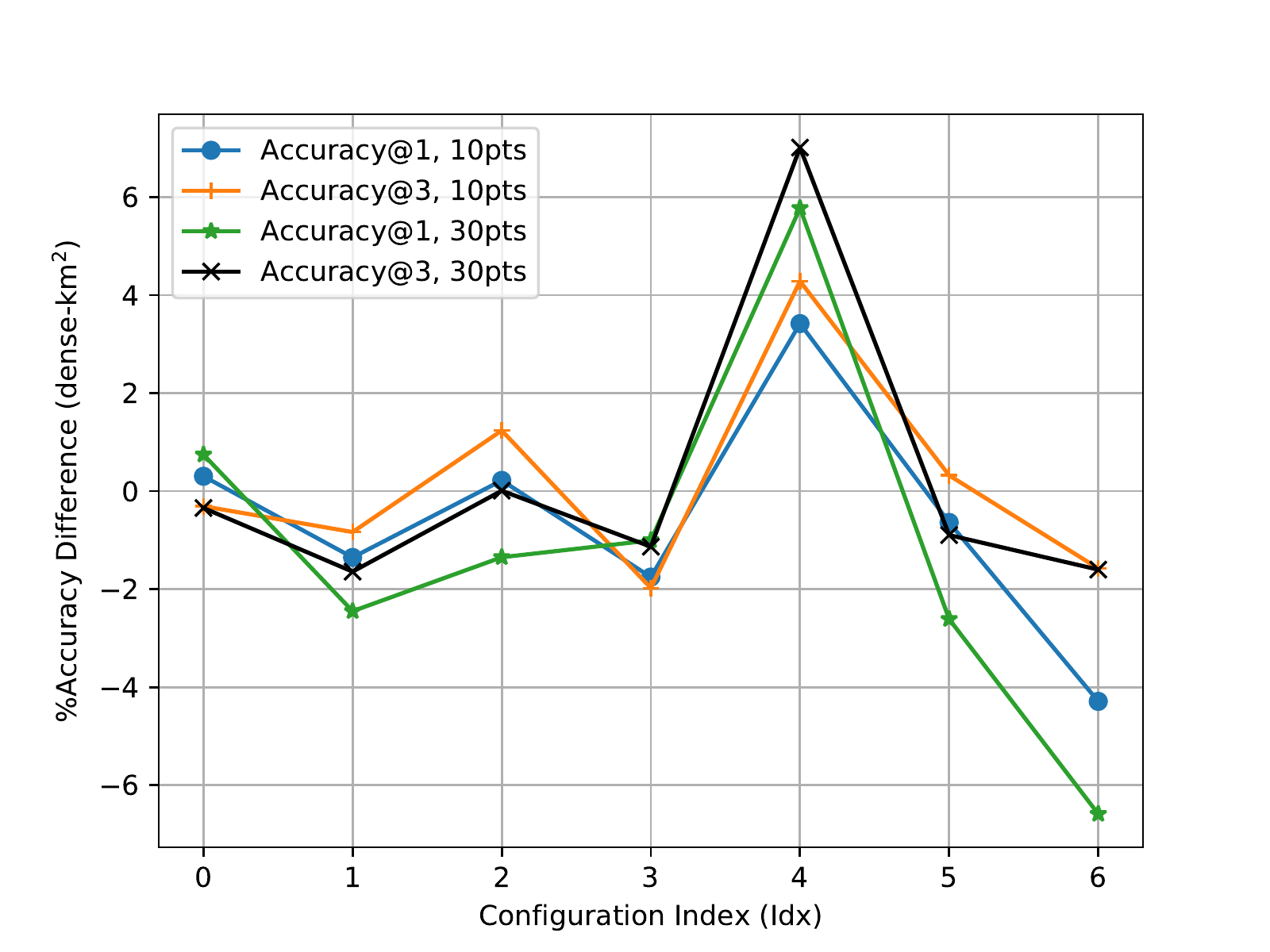}
    \caption{Difference in accuracy between the \emph{dense} and the \emph{km$^2$} cluster descriptors across dimensionality reduction configurations.}
    \label{fig:dense-vs-km}
\end{figure}

\begin{figure}[b]
    \centering
    \includegraphics[width=\textwidth]{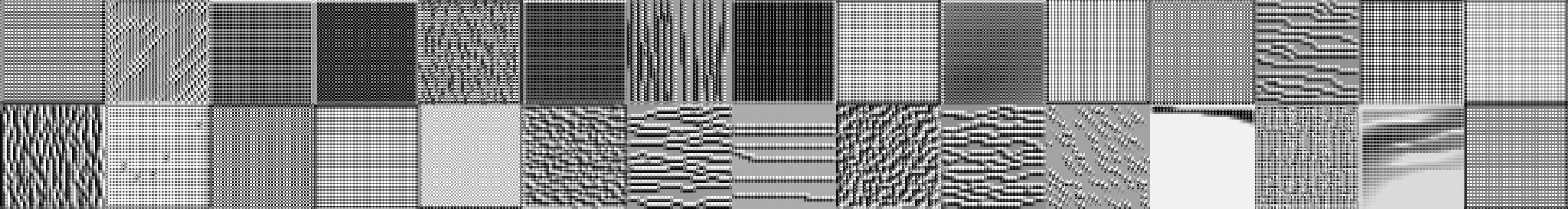}
    \includegraphics[width=\textwidth]{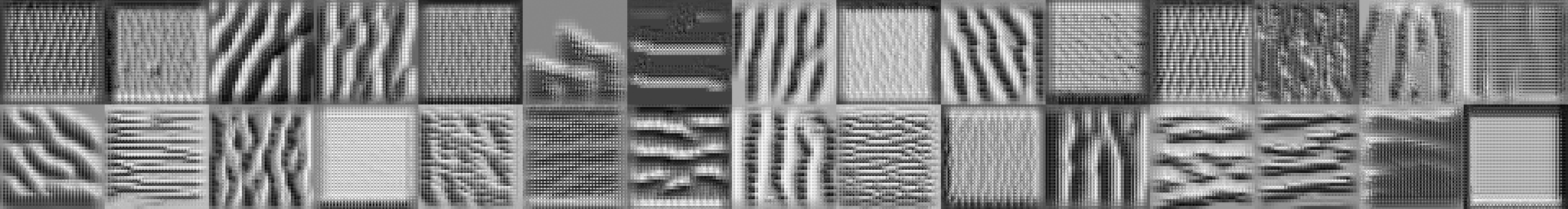}
    \caption{Convolution filters trained on Deep MC Conv A/E. The fist two rows show the first-level filters while the bottom two rows show the second-level filters.}
    \label{fig:filters}
\end{figure}

\begin{figure}
    \centering
    \begin{subfigure}{.68\textwidth}
    \centering
    \includegraphics[width=\linewidth]{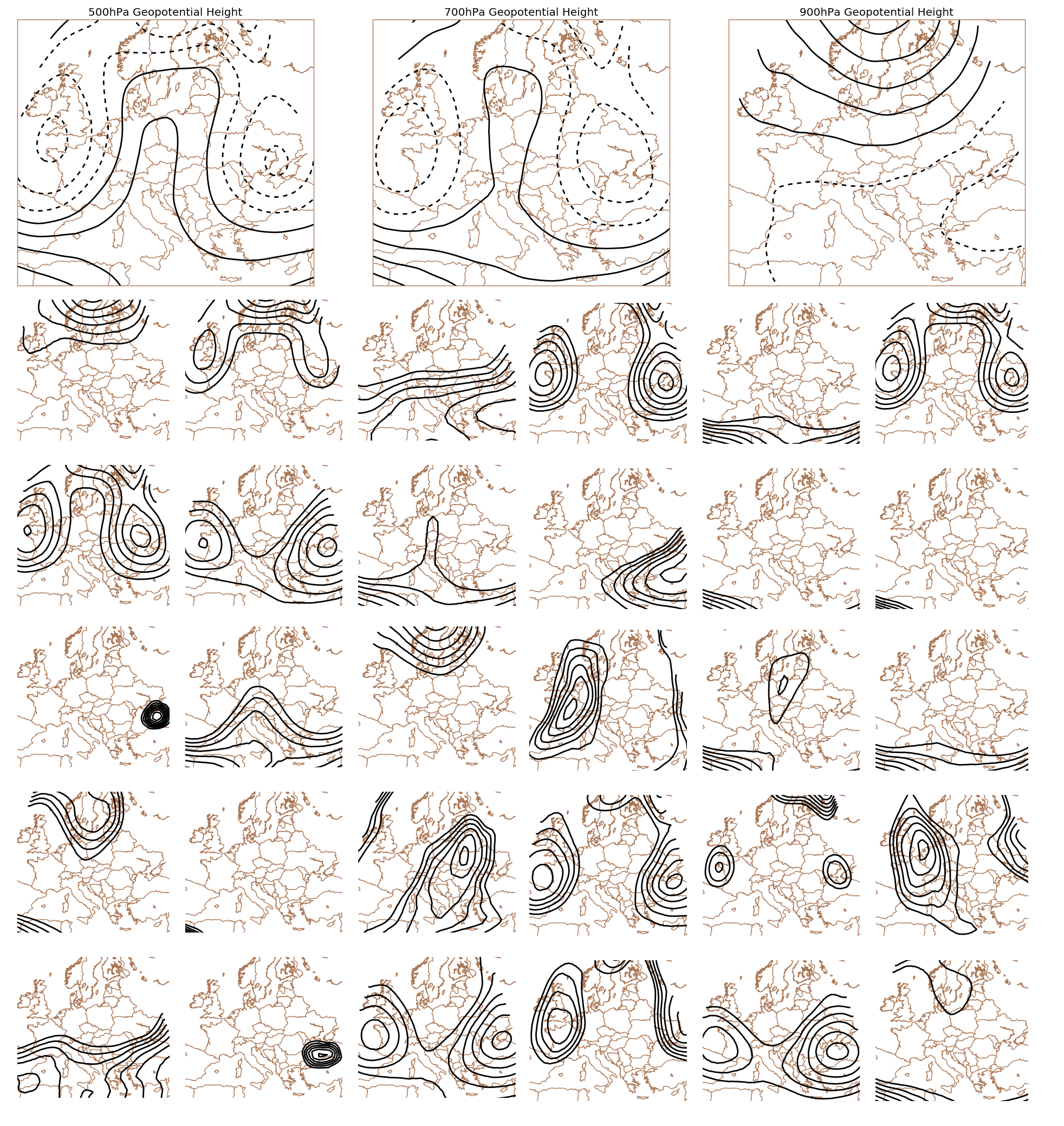}
    \caption{First layer}
    \label{fig:sub1}
    \end{subfigure}
    \begin{subfigure}{.68\textwidth}
    \centering
    \includegraphics[width=\linewidth]{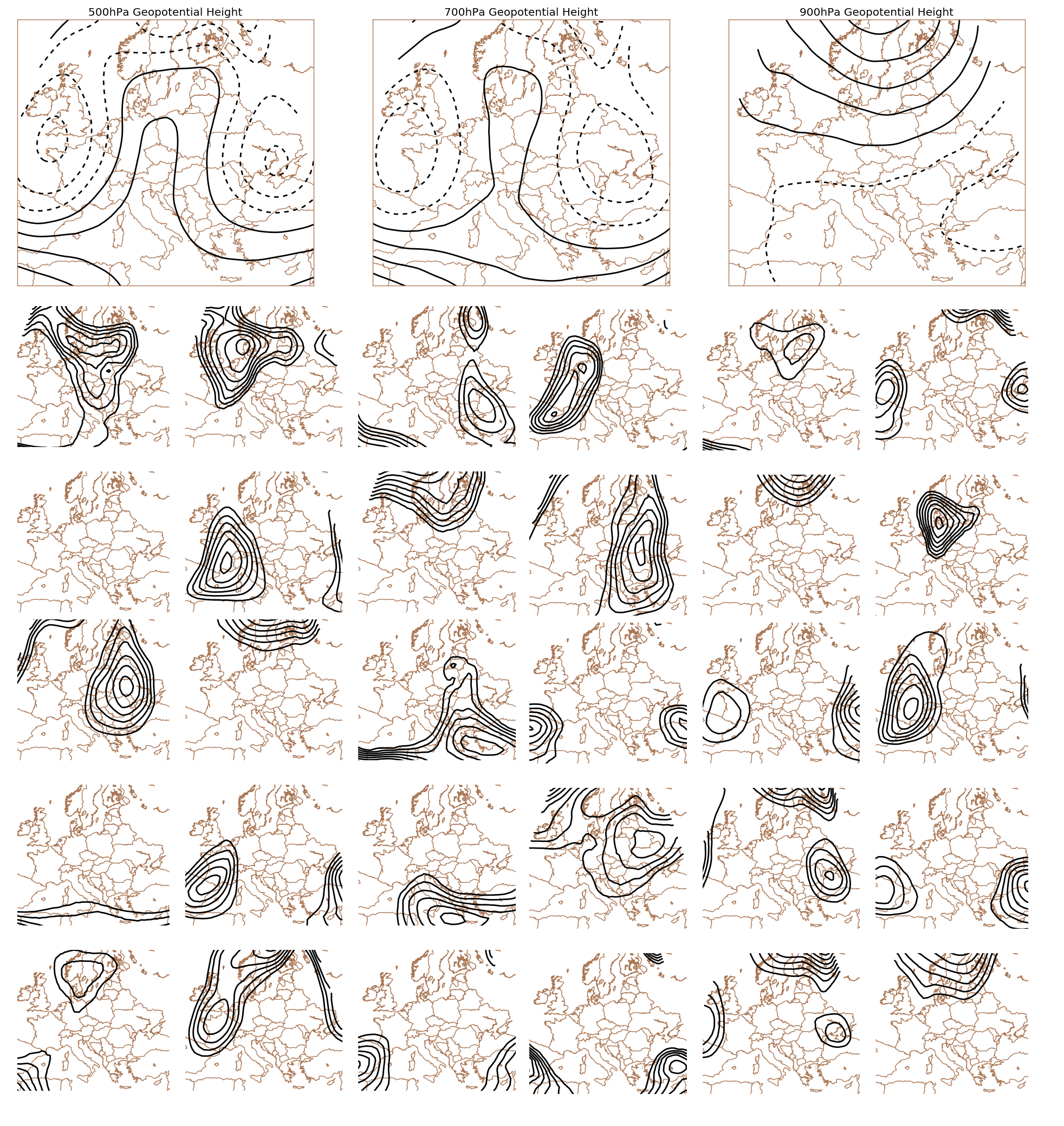}
    \caption{Second layer}
    \label{fig:sub2}
    \end{subfigure}
    \caption{Output of the first and second convolution layers in Deep MC Conv A/E for a weather sample combining data for GHT 500, 700 and 900$hPa$. Layer two exposes higher-level features, moving away from the features of the original sample.}
    \label{fig:maps}
\end{figure}

 This observation is also supported by the difference in accuracy between the two choices of cluster descriptors, shown in Fig. \ref{fig:dense-vs-km}. Here, for the most performant configuration (4, \emph{Deep MC Conv A/E}) we observe a significant and intuitive ranking in accuracy difference, with the difference increasing with the accuracy threshold (1 to 3) and the number of readings (10 to 30). This suggests that the use of multiple geopotential heights leads to more robust data representations favouring the density-based descriptor.

Fig. \ref{fig:filters} shows the input that maximises each convolution filter trained on the \emph{Deep MC Conv A/E} network \cite{Erhan2009}. Consistent with the literature, second-level filters appear to expose higher-level, more complex features of the input, which lead to increased clustering performance. The output of the first- and second-level convolution filters for a GHT sample are provided in Fig. \ref{fig:maps} highlighting the patterns discovered by each filter.

Summarising, although dimensionality reduction appears to benefit the application, and with the exception of \emph{Deep MC Conv A/E}, the depth and type of the autoencoder does not appear to be making any predictable difference on the final accuracy scores. This can be attributed to clustering errors propagated through the workflow to the final accuracy scores caused by the inherent difficulty to cluster weather, the choice of physical variables and their relationship to the dispersion as well as to the low resolution of the original data.

Having considered the above, in this application domain the timeliness of results is paramount. This, combined with visual aids (maps and dispersion patterns), user expertise in atmospheric physics and a top-3 accuracy of approximately 75\% with increasing trends with the number of readings, constitutes an arguably useful system.

\subsection{Origin Estimation as a Classification Task} 
\label{sub:origin_estimation_as_a_classification_task}
As a supplement to this study, we experimented with deep convolutional networks \cite{LeCun1998} treating the estimation of release origin as a classification problem. In this preliminary study we made use of the data produced as an evaluation dataset for the cluster-based approach, covering the years 1994 and 1995. For these two years we had previously calculated dispersions for 20 nuclear facilities in Europe and for 3-day periods. This resulted in 4480 samples which include weather and their corresponding dispersion patterns. We increased our weather and dispersion samples ten-fold. The weather dataset was increased by introducing random noise of $\pm10\%$ on GHT values, while its dispersion counterpart was increased by randomly sampling 30 points from each dispersion pattern and applying a Gaussian filter, similarly to the evaluation dataset used in the cluster-based evaluation. This resulted in a total of 49,280 samples. For evaluation, we used a separate set of 22,400 samples created from the same time period applying the same procedure.

We trained our networks to classify weather and reading patterns into one of the 20 nuclear facilities. All networks exhibit two consecutive convolution layers per input type (weather and readings/dispersion pattern) followed by maxpool layers. They are then reshaped and concatenated before they are jointly fed into a series of four fully connected layers with the last one being a softmax. The simplest network uses a single size of filters (4$\times$4 for the 64$\times$64 weather data, and 10$\times$10 for the 167$\times$167 dispersion data -- dispersion data are resized to reduce the network's memory footprint). The other two networks use two parallel convolutional threads for each input type, each with different filter sizes (1$\times$1 -- 4$\times$4 in parallel to 1$\times$1 -- 6$\times$6 for weather data and 1$\times$1 -- 10$\times$10 in parallel to 1$\times$1 -- 16$\times$16 for readings/dispersion data). The purpose of this choice was to capture spatial features at different scales. The inclusion of 1$\times$1 convolutions before larger ones was inspired by the Inception configuration \cite{inception}. The most complex network uses as input three channels of GHT at 500, 700 and 900$hPa$.
\begin{table}
\caption{Results of preliminary experiments with convolutional networks used for classification.}
\centering
{\begin{tabular}{|c||c|c|}
\hline
\emph{Name} & \emph{Accuracy@1 (\%) }&  \emph{Accuracy@3 (\%)}\\
\hline \hline
\emph{Shallow} & 58.964     & 86.352\\
\hline
\emph{Deep} & 68.104 &     90.729\\
\hline
\emph{Multi-channel} &   \textbf{75.246} &    \textbf{94.423}\\
\hline
\end{tabular}}
\label{tbl:supervised}
\end{table}

Table \ref{tbl:supervised} shows that all three classification networks outperformed the cluster-based approach significantly. Even though the evaluation task was arguably easier than the cluster-based one -- the time period was smaller and both training and evaluation sets came from the same original dataset -- the difference in performance is large enough that it warrants further investigation. A notable downside to treating this problem as a classification task is that it does not yield useful and reusable data by-products, such as weather patterns and synthetic dispersions.


\section{Application Prototype} 
\label{sec:app_prot}
Enabling decision-making in the context of high-impact events, apart from the effectiveness of the designed algorithms, depends on the visualization functionality offered, the comprehensibility of data and results and often the application responsiveness. We address the requirements of decision makers by implementing an efficient and user-friendly tool that offers visualization of the available information combining clustering by-products and dispersions.
The application prototype exploits technology integrated by the Big Data Europe integrator platform (BDI)\footnote{\url{https://www.big-data-europe.eu/platform/}}. The user interface makes use of the Sextant platform for visualising and exploring linked geospatial data\footnote{\url{http://sextant.di.uoa.gr}}.

The application prototype aids decision-making by simulating hypothetical scenarios while, if integrated with a service providing the current weather, it can also be used during real events. An event is initiated by the presence of radioactivity at certain geographical locations. By entering the marking mode, users can mark the locations of such readings. The marking mode allows the user to freely place the detection reading by hand (longitude and latitude of the mouse pointer is always shown to the user). This mode offers the ability to mimic the location of detection stations as well as the location of portable radiation detection devices.
To estimate the origin's location the application needs to know, apart from the detection readings, the state of the current weather.
The current weather is visualised as arrows by combining the west-east and north-south wind magnitudes. Currently, wind direction at three pressure levels can be displayed (500, 700 and 900$hPa$).

The concentration of radioactivity based on the clustering descriptors has been precomputed for two radionuclides typically emitted during nuclear accidents: Caesium-137 (Cs-137) and Iodine-131 (I-131). Pollutants to be simulated as well as the source estimation method (various clustering methods) are part of the prototype's configuration parameters and can be selected via the control panel. 

The prototype offers fast response times due to the methodology adopted and its implementation using performant big data technologies. Efficient and reliable access to neural network models and weather data as well as caching for visualisation are accommodated by the BDI platform.

\section{Conclusions and Future Work} 
\label{sec:conclusions}
In this work we presented a novel machine-learning-driven, cluster-based approach for rapid origin estimation during nuclear incidents. Our approach is inspired by image processing, deep learning and data mining, extending useful techniques to weather and atmospheric dispersion data. Incentivised by the need to create ML-ready reusable data products to enable the creation of sophisticated applications, we discussed the derivation of usable weather patterns via clustering. We employed deep autoencoders to derive robust latent features on which to apply clustering. Our evaluation shows that deep feature extraction followed by weather clustering can improve response efficiency during nuclear emergencies considerably, complementing existing operational systems.

Directions for future research and development may include the following: 
(1) evaluation of alternative classification approaches that learn origins, similar to the ones presented here, but also dispersion patterns 
(2) the formulation of novel clustering algorithms that take into account features such as periodicity and temporal density for cluster formation; 
(3) the inclusion of additional atmospheric dispersion models and learning model ensembles; 
(4) evaluation of alternative learning algorithms, such as recurrent neural networks (RNNs), to explicitly account for the temporal component of such analyses; 
(5) extension of the application to include all nuclear facilities in Europe and further to consider all possible source locations (\ie{} all grid cells as potential sources), to be applicable in all radiological incidents. This would enable the identification of arbitrary sources and would be applicable to scenarios similar to the recent discovery of Ru-106 in Europe; and 
(6) the evolution of the application to an operational system that learns continuously with new meteorological data.

\section{Software Availability and Acknowledgements}
\label{sec:ack}
The source code for the experiments and evaluation described in Section \ref{sec:results_and_discussion} are available at \url{https://github.com/davidath/ncsr-atmo-learn}.
The source code and installation instructions for the prototype described in Section \ref{sec:app_prot} can be found at \url{https://github.com/iaklampanos/bde-pilot-2}. A video demonstration of the prototype, also including extensions, such as the drawing and selection of known fixed radioactivity detectors and the depiction of inhabited locations below the estimated plume, can be found at \url{https://vimeo.com/227245883}.

This work has received funding from the European Union's
Horizon~2020 research and innovation programme under grant agreement
No~644564.
For more details please visit \url{https://www.big-data-europe.eu}.


\section*{References}
\bibliographystyle{elsarticle-num}
\bibliography{sc5_inv_bib}


\end{document}